# Topic Modeling in Marathi

Sanket Shinde
*Department Of Electronics and Technology,*
*Vishwakarma Institute Of Technology, Pune*
sanket.shinde19@vit.edu

Raviraj Joshi
*L3Cube Labs, Pune*
*Indian Institute Of Technology Madras, Chennai*
ravirajoshi@gmail.com

*Abstract*—While topic modeling in English has become a prevalent and well-explored area, venturing into topic modeling for Indic languages remains relatively rare. The limited availability of resources, diverse linguistic structures, and unique challenges posed by Indic languages contribute to the scarcity of research and applications in this domain. Despite the growing interest in natural language processing and machine learning, there exists a noticeable gap in the comprehensive exploration of topic modeling methodologies tailored specifically for languages such as Hindi, Marathi, Tamil, and others. In this paper, we examine several topic modeling approaches applied to the Marathi language. Specifically, we compare various BERT and non-BERT approaches, including multilingual and monolingual BERT models, using topic coherence and topic diversity as evaluation metrics. Our analysis provides insights into the performance of these approaches for Marathi language topic modeling. The key finding of the paper is that BERTopic, when combined with BERT models trained on Indic languages, outperforms LDA in terms of topic modeling performance.

## I. INTRODUCTION

As the volume of digital textual data grows exponentially, manually analyzing and extracting insights from these datasets becomes increasingly challenging. This is where topic modeling emerges as a powerful tool for automated text analysis and summarization. The primary goal of topic modeling is to uncover the latent topics within a corpus of text and represent them in a way that is interpretable by humans. These algorithms operate on the assumption that each document in a corpus can be expressed as a mixture of topics and that each topic can be represented as a distribution of words. By analyzing the co-occurrence patterns of words across multiple documents, topic modeling algorithms can identify the most probable topics present in the corpus.

The ability to identify latent themes in large collections of text has diverse applications across industries. For instance, social media companies can use topic modeling to analyze user sentiment and opinions, while market research firms can explore customer feedback to detect recurring themes and pain points [1]. In academia, topic modeling serves as a valuable tool for identifying emerging trends and research areas within various fields [2].

Despite its benefits, topic modeling faces significant challenges, particularly for low-resource languages (LRLs). The vast majority of today's natural language processing (NLP) research focuses on just 20 of the world's 7,000 languages, leaving most languages underexplored [3]. These low-resource languages, which include approximately 2,000 spoken across Africa and India, represent a population exceeding 2.5 billion people. Developing technologies for LRLs has immense economic and social potential.

A major challenge in applying topic modeling to LRLs lies in their limited linguistic resources. These algorithms depend heavily on large corpora of text to achieve accurate results, which is often unattainable for languages with restricted datasets. Addressing these limitations is critical to ensuring that NLP tools like topic modeling are accessible and effective across the linguistic diversity of our world.

This paper focuses on the Marathi language, a low-resource language in the context of natural language processing (NLP). Marathi is an Indo-Aryan language spoken primarily in the Indian state of Maharashtra and is the fourth most spoken language in India. Despite its widespread use, Marathi remains relatively understudied in NLP, with a limited amount of digital text data available for analysis. This scarcity of resources poses challenges for tasks such as language modeling, sentiment analysis, and topic modeling. However, recent advancements in transfer learning and unsupervised learning techniques offer promising avenues for building accurate language models and conducting effective NLP tasks for Marathi.

To address these challenges, we explored two prominent topic modeling methods—BERTopic and Latent Dirichlet Allocation (LDA)—and evaluated their performance on Marathi text data. BERTopic leverages BERT embeddings to identify topics in text corpora, while LDA, a well-established probabilistic model, has long been a standard for topic modeling in NLP. Our study demonstrates that BERTopic, powered by advanced sentence embeddings, provides meaningful insights into the themes and cultural nuances present in Marathi text.

Using SentenceTransformer, we prepared sentence embeddings for Marathi text data, enabling dense vector representations of sentences. This method proved effective in encoding Marathi sentences, laying a robust foundation for downstream NLP tasks. To assess the relative performance of BERTopic and LDA, we compared their topic coherence scores and analyzed the most frequent words in each identified topic. Our results show that while both methods are effective for topic modeling in Marathi, BERTopic consistently outperforms LDA in terms of topic coherence, highlighting its suitability for low-resource languages like Marathi.

Additionally, we utilized BERT models from L3Cube [4], such as marathi-bert-v2, and Sentence-BERT (SBERT) models, including marathi-sentence-similarity-sbert, marathi-sentence-bert-nli, indic-sentence-similarity-sbert, and indic-sentence-bert-nli. These models were instrumental in building and analyzing topic models tailored to Marathi datasets of varying lengths, ranging from short to long texts.

Our research revolves around the creation and evaluation of topic models for Marathi, emphasizing the advantages of advanced embeddings in capturing thematic structures. By applying BERTopic and LDA to diverse datasets, we provided an in-depth analysis of their strengths and limitations. This study contributes to a growing body of work on NLP for low-resource languages, offering a foundation for future research in Marathi text analysis and advancing the broader understanding of topic modeling techniques.

## II. Literature Survey

Topic Modeling has emerged as a powerful tool for uncovering latent thematic structures in large text corpora. While widely applied in English and other major languages, its application to Marathi, a morphologically complex Indo-Aryan language with over 80 million speakers, poses unique challenges and necessitates exploration of adapted techniques. This section reviews existing works on topic modeling in Marathi, highlighting their insights and limitations, before situating our proposed approach within this landscape.

Various methodologies have been developed to extract meaningful topics from textual data, each with its unique strengths and applications. [5] A systematic literature review identified several topic modeling techniques such as Latent Dirichlet Allocation (LDA), Latent Semantic Analysis (LSA), Support Vector Machine (SVM), Bi-term Topic Modeling (BTM).

[6] LDA depicts topics through word probabilities, where the words with the highest probabilities within each topic generally provide a clear indication of the topic. Examining the word probabilities derived from LDA often offers valuable insights into the nature of a given topic.

[7] Maarten Gr. presents BERTopic, a novel topic model that creates coherent topic representations by utilizing a class-based variant of TF-IDF and clustering algorithms. In order to extract document-level information, BERTopic first builds document embeddings using a language model that has already been trained. Next, semantically comparable clusters of documents representing different topics are created by reducing the dimensionality of the document embeddings. To extract the topic representation from each subject, a class-based variant of TF-IDF is used.

[8] employed Word2Vec to generate a topic model on Hindi corpus. The model was evaluated on two corpuses, Hindi and English mixed corpus and a pure Hindi corpus. According to their findings, word2vec combined with clustering algorithms creates a powerful topic modeling and detector for unstructured text in Hindi, English, and mixed corpora.

[9] discusses the challenges to topic modeling in Hindi language. Challenges like no-capitalization, lack of uniformity in writing styles, expressions with multiple words and lack of NLP resources are discussed. These challenges apply for Marathi language too as Marathi is also written using devanagari script. Similar difficulties can also be encountered with other Indian languages.

The text [10] details the process of topic detection using Latent Dirichlet Allocation (LDA) on a Marathi corpus. The corpus, consisting of Marathi stories and poems, is pre-processed and converted into a vector space model (VSM) using TF-IDF. LDA is then applied to the VSM to identify coherent topics. The number of topics is determined based on the coherence score, and the model's performance is evaluated using intrinsic and extrinsic metrics.

There are many approaches to evaluate topic models, including quantitative metrics such as coherence score, perplexity, and topic diversity. While these metrics can provide valuable insights into the quality of a topic model, they have their limitations and cannot fully capture the nuances and complexity of language that humans can perceive. [11] introduces a range of evaluation methods, such as estimating normalizing constants and document completion, which were applied to assess the performance of Latent Dirichlet Allocation (LDA).

## III. Methodology

This study examines different topic modeling methodologies for the Marathi language and evaluates their performance based on topic coherence.

In Python, there are several libraries that provide robust and efficient tools for implementing Latent Dirichlet Allocation (LDA) in topic modeling. In this study, we utilized established Python libraries such as Gensim to perform the analysis based on Latent Dirichlet Allocation (LDA).

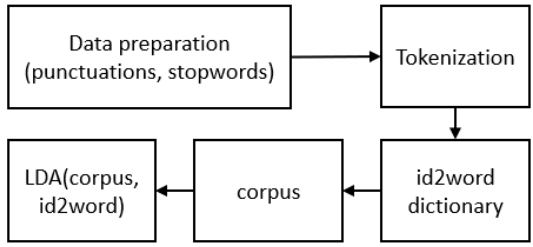

Fig. 1. Block Diagram for Gensim's LDA

Fig 1 shows the system diagram for LDA. The initial phase of our approach involved data preparation, where we systematically removed punctuations, stop words, and non-Devanagari words. Subsequently, we performed tokenization to break down the text into individual units. Moving forward, we focused on creating a dictionary to establish a mapping between unique words in the corpus and their corresponding integer IDs. This dictionary, referred to as id2word, plays a pivotal role in Gensim's LDA model. It not only facilitates efficient text data processing but also allows the model to associate each word in the documents with a distinct identifier.

Following the establishment of the id2word dictionary, the subsequent step entailed transforming the tokenized documents into a bag-of-words representation. In this representation, each document underwent a conversion into a sparse vector, where each element denoted the frequency of a unique word within the document. With the completion of the id2word and bag-of-words steps, our prepared data, in the form of the id2word dictionary and the corpus representation, was seamlessly integrated into the LDA model implemented using Gensim. This pivotal step marked the transition from data preprocessing to the application of the LDA model, allowing for the extraction of latent topics and insights from the input corpus.

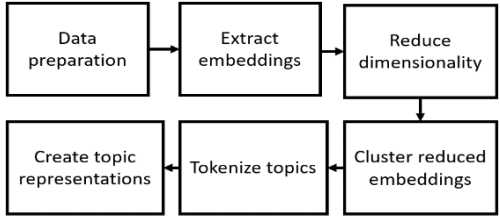

Fig. 2. Block Diagram for BERTopic

BERTopic employs a multi-step process to derive coherent topic representations. Initially, documents are converted into numerical embeddings using pre-trained language models, typically from sentence-transformers. To address the challenge of high dimensionality, UMAP is applied to reduce the embedding space while preserving essential data structure. Subsequently, HDBSCAN clusters the reduced embeddings into groups of semantically similar documents, accommodating clusters of varying shapes and densities.

A bag-of-words representation is then constructed for each cluster by combining all its documents into a single document. This representation serves as the foundation for calculating topic representations. By applying a modified TF-IDF approach, known as c-TF-IDF, the algorithm determines the significance of words within each cluster, effectively capturing the core terms of the topic. Optionally, BERTopic allows for fine-tuning topic representations using advanced language models like GPT or T5, potentially enhancing topic quality and accuracy.

Here, we used a variety of bert and non-bert models like [12] MahaSBERT-STS, MahaSBERT, MahaBERT-V2, [13] IndicSBERT, IndicSBERT-STS and Muril to generate embeddings.

In [7] Maarten Gr. uses topic coherence and topic diversity as two metrics to compare the BERTopic models and we also use these same metrics to compare our bert models. Topic coherence basically tells us how related the words are in a given topic. And in simple terms, topic diversity refers to how different or varied the topics are within a set of documents or discussions.

When topics are shown to users, each topic t is typically shown as a list of the M = 5,..., 20 most likely words for that subject, arranged in descending order based on their probabilities unique to that topic.

IV. RESULTS

We have compared the performances of 5 bert models, such as MahaSBERT-STS, MahaSBERT, MahaBERT-V2, IndicSBERT, and Muril. A detailed comparison of the topic coherence shown by these models can be found in Table 1.

|  | MahaSBERT-STS | *MahaSBERT* | MahaBERT-V2 | IndicSBERT | IndicSBERT-STS | MURIL |
|---|---|---|---|---|---|---|
| LPC Dataset | 0.70 | 0.63 | 0.71 | 0.72 | 0.72 | 0.65 |
| LDC Dataset | 0.78 | 0.71 | 0.82 | 0.74 | 0.79 | 0.80 |
| SHC Dataset | 0.79 | 0.72 | 0.73 | 0.77 | 0.81 | 0.67 |

Table. 1. Topic Coherence of different BERT models against the LPC, LDC and SHC datasets using BERTopic

It was observed that BERTopic combined with bert models trained on Indic languages performs better than LDA. In BERTopic, monolingual bert model MahaBERT-V2 performs consistently better than other bert models in the 3 datasets.

[4] MahaBERT-V2 is outperforming MuRIL in our evaluations, demonstrating superior performance across the tasks. When it comes to document classification, [12] SBERT

models are particularly effective on the SHC dataset, which consists of shorter documents, as they are better at capturing the relevant features and context within brief texts. Conversely, BERT models show greater effectiveness on the LDC dataset, where their ability to process and understand longer and more complex documents gives them an advantage. For the LPC dataset, which consists of medium-length documents, the performance of SBERT and BERT models is closely matched, with neither model showing a significant advantage over the other.

To further increase the accuracy of the BERTopic models, the data was pre-processed by removing Marathi stop words, URLs and non-devanagari words.

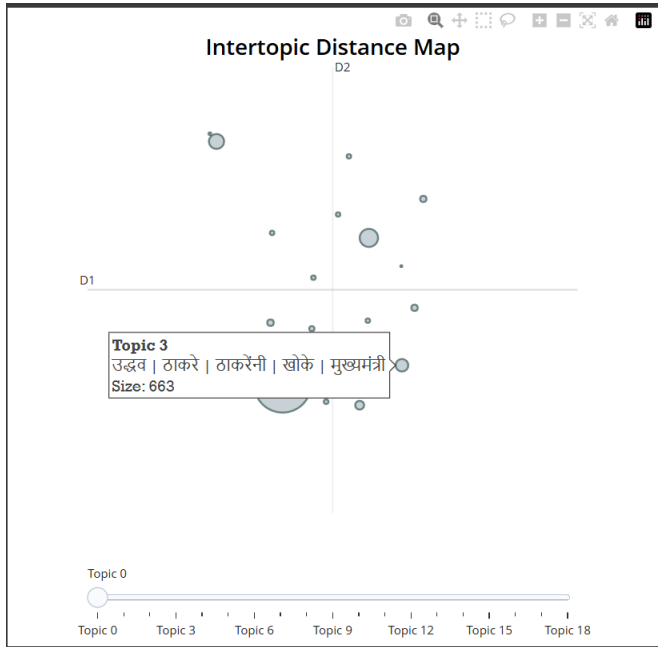

Fig. 3.Visualization of the similar words belonging to a similar topic

As shown in Fig 3. the topics in BERTopic are represented as a collection of similar words.

|  | Latent Dirichlet Allocation |
|---|---|
| LPC Dataset | 0.34 |
| LDC Dataset | 0.55 |
| SHC Dataset | 0.45 |

Table. 2. Topic Coherence of LDA model across different datasets

The topic coherence score of the LDA model across different datasets were lesser than BERTopic, and the accuracy of words for certain topics were also lesser.

## V. CONCLUSION

In this paper, we evaluated various topic modeling techniques and applied them to a Marathi dataset. We explored several methods used for topic modeling, including Latent Dirichlet Allocation (LDA) and BERTopic. We trained the BERTopic model on three different datasets: LDC (long documents), LPC (medium documents), and SHC (short documents). Furthermore, in BERTopic, we used both multilingual and mono-lingual BERT models trained on Marathi datasets. The coherence score was used as a metric to compare the performance of the models. We observed that BERTopic models performed better than LDA models with a higher coherence score. The comparison highlights a potential advantage of BERTopic over LDA in terms of generating more coherent and accurate topics. This could be attributed to factors such as the use of pre-trained language models in BERTopic, which can capture semantic relationships between words more effectively than LDA's statistical approach.